\documentclass{ecai}
\usepackage{graphicx}
\usepackage{latexsym}
\usepackage{color}
\usepackage{hyperref}

\usepackage{algpseudocode}
\usepackage{algorithm}
\usepackage{algorithmicx}
\usepackage{amsmath,amsfonts,bbm}
\usepackage{mathtools}
\usepackage{tabularx}
\usepackage{graphicx}
\usepackage{subcaption} 
\usepackage{comment}
\usepackage{soul}
\usepackage{booktabs}
\definecolor{corlinks}{RGB}{0,0,170}
\definecolor{cormenu}{RGB}{0,0,170}
\definecolor{corurl}{RGB}{0,0,170}
\definecolor{papercolor}{rgb}{0.82, 0.1, 0.26}

\definecolor{corm}{RGB}{0,0,130}

\hypersetup{
colorlinks=true,
urlcolor=corlinks,
linkcolor=corlinks,
menucolor=cormenu,
citecolor=corlinks,
pdfborder= 0 0 0
}

\newcommand{\be}{\begin{equation}}
\newcommand{\ee}{\end{equation}}
\newtheorem{corollary}{Corollary}[theorem]
\newcommand{\mc}{\mathcal}

\begin{document}

\begin{frontmatter}

\title{Generalizing similarity in noisy setups: the DIBS phenomenon}

%___________________________________

\author[A]{\fnms{Nayara}~\snm{Fonseca}\footnote{Equal contribution.  Emails: nayara.fonsecadesa@physics.ox.ac.uk; veronica.guidetti@unimore.it.}$^{*,}$}
\author[B]{\fnms{Veronica}~\snm{Guidetti}$^{*,}$}

\address[A]{Rudolf Peierls Centre for Theoretical Physics, University of Oxford, Oxford, UK}
\address[B]{UNIMORE, FIM Department, Via G. Campi 213/a, 41125, Modena, Italy}

\begin{abstract}
This work uncovers an interplay among data density, noise, and the generalization ability in similarity learning. We consider Siamese Neural Networks (SNNs), which are the basic form of contrastive learning, and explore two types of noise that can impact SNNs, Pair Label Noise (PLN) and Single Label Noise (SLN). Our investigation reveals that SNNs exhibit double descent behaviour regardless of the training setup and that it is further exacerbated by noise.  We demonstrate that the density of data pairs is crucial for generalization. When SNNs are trained on sparse datasets with the same amount of PLN or SLN, they exhibit comparable generalization properties.  However, when using dense datasets,  PLN cases generalize worse than SLN ones in the overparametrized region, leading to a phenomenon we call Density-Induced Break of Similarity (DIBS). In this regime, PLN similarity violation becomes macroscopical, corrupting the dataset to the point where complete interpolation cannot be achieved, regardless of the number of model parameters. Our analysis also delves into the correspondence between online optimization and offline generalization in similarity learning. The results show that this equivalence fails in the presence of label noise in all the scenarios considered.
\end{abstract}

\end{frontmatter}

\section{Introduction}
\label{sec:introduction}

In recent years, several works have studied generalization in neural networks (NNs) and the connection between the classical underparametrized regime, where the number of training samples is larger than the number of parameters in the model, and that of deep learning, where the opposite is usually the norm. Indeed, the empirical success of overparametrized NNs challenges conventional wisdom in classical statistical learning. It is widely known among practitioners that larger models (with more parameters) often obtain better generalization performance \cite{szegedy2015going, huang2019gpipe, radford2019language}.

Two frameworks adopted to study generalization in regression or classification tasks are \emph{Double Descent} (DD) and \emph{online/offline learning correspondence}, which we describe in the following.  
The DD from \cite{Belkin:2019} connects ``classical'' and ``modern'' machine learning by observing that once the model complexity is large enough to interpolate the dataset (i.e., when the training error reaches zero), the test error decreases again, reducing the generalization gap. This pattern has been empirically observed for several models and datasets, ranging from linear models, as in \cite{loog2020brief}, to modern deep neural networks, as in \cite{spigler2019jamming, NakkiranDeepDD:2020}.
Instead, the online/offline learning correspondence of \cite{Nakkiran:2020} studies the relationship between online optimization and offline generalization. The conjecture, empirically verified on supervised image classification, states that generalization in an offline setting can be effectively reduced to an optimization problem in the infinite-data limit. This means that online and offline test errors coincide if the NN is trained for a fixed number of weight updates. This setup aims to \emph{link} the under- and overparametrized models: the infinite-data limit (online) sits in the under-parameterized region (number of samples $>$ number  of parameters), while the finite-data case (offline) corresponds to the overparametrized regime  (number of samples $<$ number  of parameters). Here, we test if this correspondence is also valid for similarity tasks.

\emph{DD phenomenon and online/offline correspondence are two complementary approaches that look at different generalization properties:} while the DD analysis studies how the network adjusts to an increasing number of parameters, the online/offline training compares the network performance by varying the dataset size while fixing the number of weight updates. Although these approaches have mainly been applied to classification and regression, if they are associated with some fundamental properties of deep neural networks, they should also hold for other tasks such as similarity learning.

There are key differences between similar-different discrimination and classification. For similarity learning, the relation among features is crucial, but not necessarily the features themselves.  For this reason, a priori, it is not possible to predict whether the DD behavior and the online/offline learning correspondence will also occur in similarity problems.
To take the first steps towards understanding how deep neural networks generalize in similarity learning, we export both frameworks to the simplest contrastive learning representative, Siamese Neural Networks (SNNs) from \cite{siamese:1993, Chopa2005}. 
A Siamese architecture is made of two identical networks sharing weights and biases that are simultaneously updated during supervised training. The two networks are connected by a final layer, which computes the distance between branch outputs. SNNs are trained using pairs of data that are labeled as similar or different. The task of a successfully trained network is to decide if the pair samples belong to the same class.

\begin{figure*}[!ht]
\center
 \begin{sc}
    \includegraphics[width=0.7\textwidth]{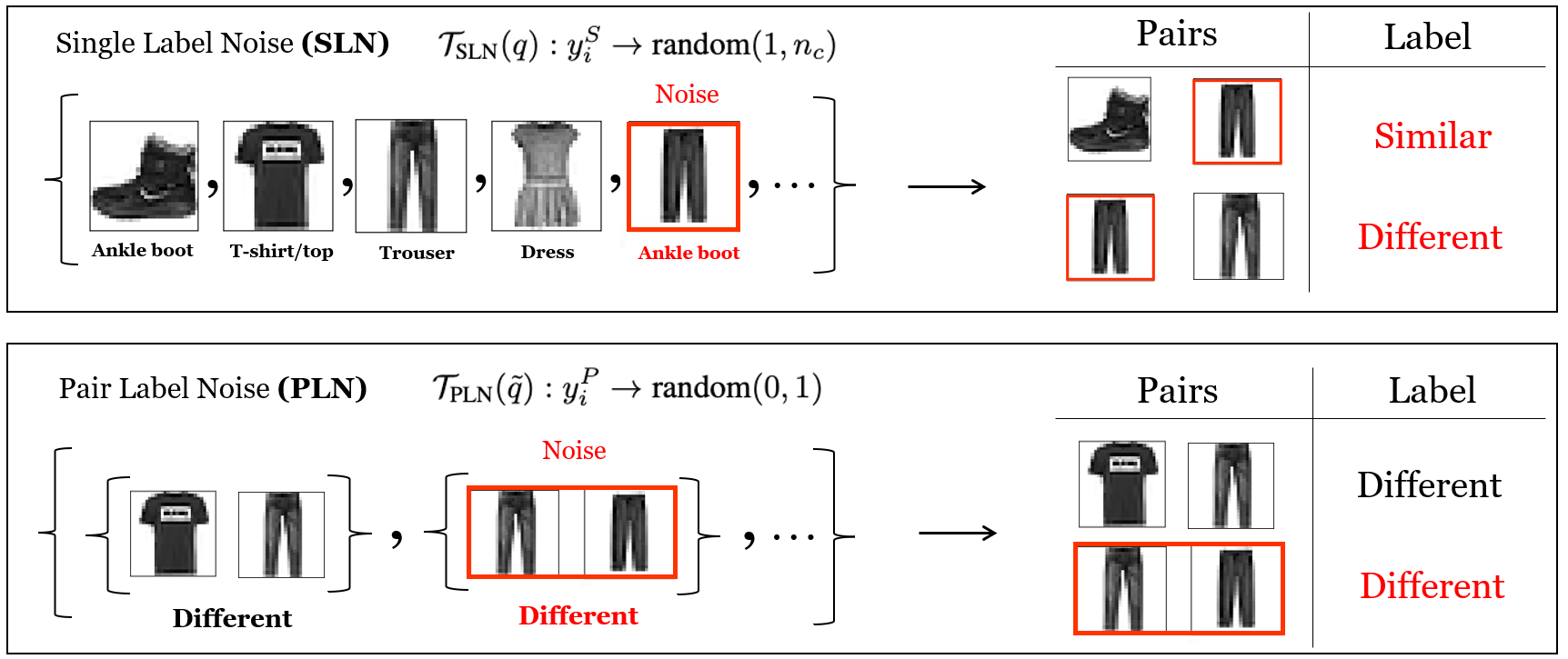} \\~ \\[-5pt]
    \begin{tabular}{  >{\centering\arraybackslash} m{8cm}>{\centering\arraybackslash} m{8cm}}
    \small{Sparse (S1)} & \small{Dense (S2)} \\
    \includegraphics[height=0.18\textwidth]{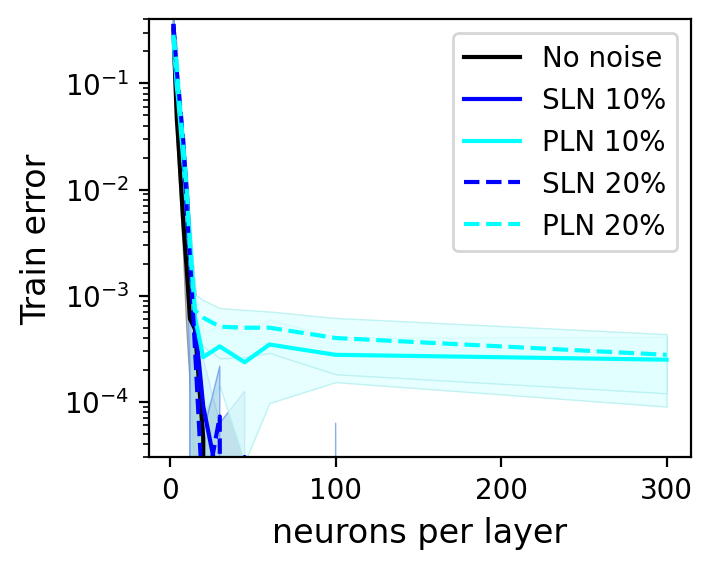}  \includegraphics[height=0.18\textwidth]{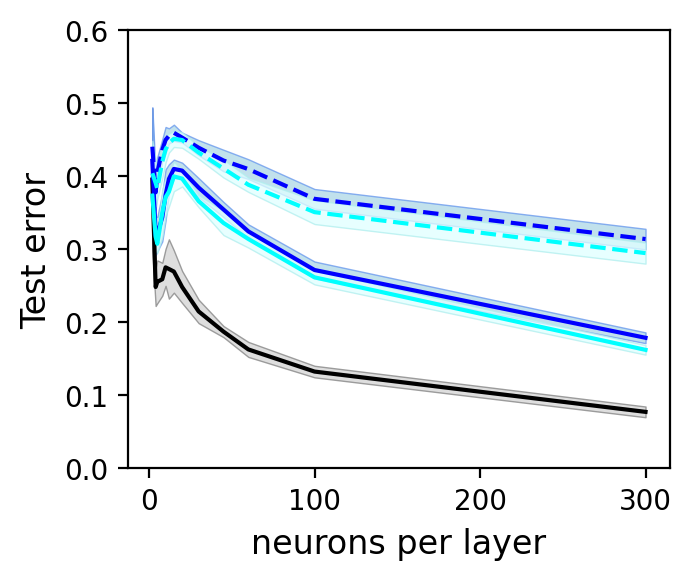} &  \includegraphics[height=0.18\textwidth]{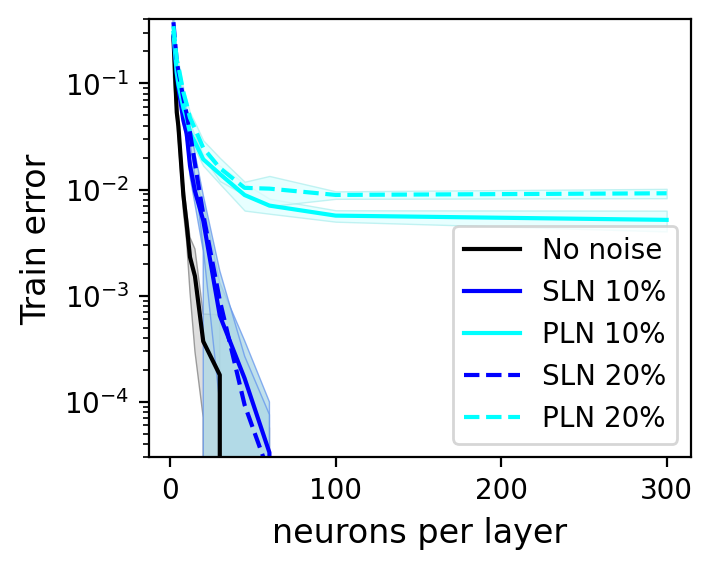}  \includegraphics[height=0.18\textwidth]{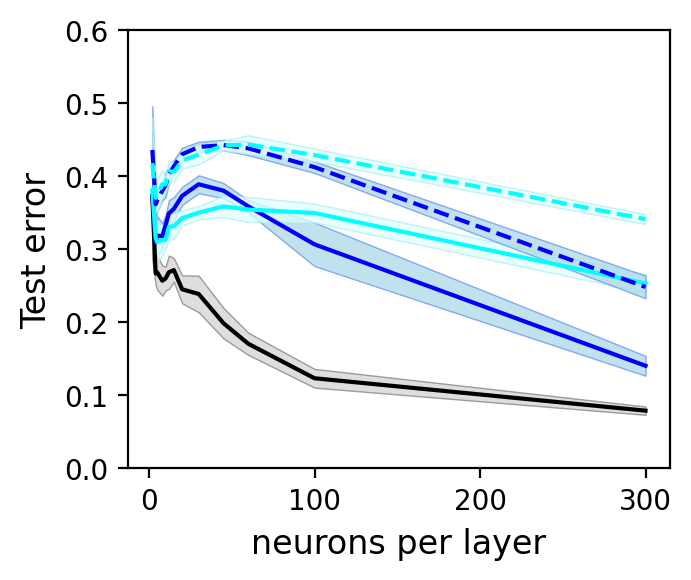} \\[0pt]
    \end{tabular}
\end{sc}
  \caption{\textbf{Top:} Illustration  of SLN and PLN applied to image samples and the resulting dataset of pairs. As discussed in the text, PLN leads to  inconsistent relations in the dataset. This effect becomes more apparent when using dense datasets. On the other hand, similarity breaking does not appear in SLN, where the similarity relations may go against image features but are self-consistent.
 \textbf{Bottom:} Train  and test  errors  as a function of model size for sparse (S1) and dense (S2) configurations. We consider a 3-layer MLP with ReLU activation functions trained on sparse and dense pairs of MNIST with 10\% and 20\% effective noise (see Sec.\,\ref{sec:experiments} for details). Note that both no-noise and SLN cases reach complete interpolation in the training set, while PLN train error no longer vanishes by increasing the number of network parameters. \label{fig:PLN_SLN}  \vspace{-0.1cm}}
\end{figure*}

Studying the DD and online/offline correspondence in SNNs and comparing the results with those found in classification problems requires identifying which properties/characteristics of the training set most influence similarity learning. We identified two crucial sources of variability: \emph{(i)} the effect of noisy data in SNNs, and \emph{(ii)} the density of pairs in the training  set. 
Noise is crucial in understanding generalization as it appears in every real-world dataset and may compromise model performance. While DD was also studied in the presence of noise,\footnote{Notably, it is  known that the DD curve is exacerbated in the presence of random label noise in supervised classification  (see, e.g., \cite{NakkiranDeepDD:2020}).}  very little (if none) attention was devoted to noise in the online/offline setting. By construction, SNNs can be affected by more complex types of noise than classification problems. This derives from the use of pairwise relations defining a similarity graph. 
To show the reaction of SNNs to different noise sources, we introduce two representative examples with distinctive properties: Single Label Noise (SLN) and Pair Label Noise (PLN), which we  describe in detail in Sec. \ref{sec:contrastive}. As we will show, SLN breaks similar/different pairs balancing but preserves similarity relations. Instead, PLN acts symmetrically on pair labels, but it breaks transitivity and, thus, similarity. See top of Fig.\,\ref{fig:PLN_SLN} for a pictorial view of SLN and PLN.
Furthermore, we show that similarity learning is strongly influenced by the \emph{density of pairs in the training set}. Concretely, we demonstrate how pairs created from populations with different levels of similarity graph density (or image diversity) give rise to very different learning models. This means that the average number of different images appearing in a set of pairs can significantly impact the learning process and ultimately influence the performance of the model. We discuss sparse and dense connections in detail in Sec.\,\ref{sec:contrastive}
\vspace{0.1cm}

The results in this work are summarized as follows. 
\begin{itemize}
    \item  DD  clearly appears in SNNs, regardless of the noise level, a phenomenon rarely found in classification problems in the absence of noise.
    \item DD is exacerbated by noise (in line with \cite{NakkiranDeepDD:2020}), and its shape is affected by the density of pairs in the training set. While SNNs trained on sparse datasets show similar DD curves in the presence of SLN and PLN, these become quite distinct when the similarity relations in the training set are dense. Specifically, the interpolation threshold in the presence of PLN requires more parameters, and its test error remains higher in the overparametrized region. An example of this behavior is shown at the bottom of Fig.\,\ref{fig:PLN_SLN}. 
    \item 
    We show that complete interpolation  (training error $= 0$) cannot be achieved in the PLN scenario with dense connections and derive an upper and lower bound for the asymptotic training error value in the deep overparametrized regime. We call this phenomenon \emph{Density-Induced Break of Similarity (DIBS)}.
    \item We test the correspondence between offline generalization and online optimization for similarity learning. We study how the architecture and the presence of noisy labels can differently impact these two regimes. We find that the conjecture only holds for clean data.
  \item In the presence of label noise, we find that the online/offline correspondence breaks down for all choices of training settings considered. In particular, the effect of label noise is  notably more relevant in the offline case.

\end{itemize}

\vspace{-0.05cm}

\subsection{Related work}
\label{sec:related}

Over the past few years, significant strides have been made to understand how neural networks generalize in the presence of noise in classification problems (e.g., \cite{class_label_noise:2018, deep_class_noise:2019, unsupervised_class_noise:2019, info_class_noise:2020, survey_class_noise:2020}). Remarkably, the DD phenomenon enabled a closer examination of the NN behavior as the number of trainable parameters, the evolution time, and the size of the dataset vary \cite{NakkiranDeepDD:2020,bodin2021model,heckel2020early,pezeshki2021multi}. Subsequently, other works have produced analytical studies of some of these phenomena \cite{d2020double, d2020triple, mei2022generalization}. Another complementary tool used to study generalization in classification tasks is the online/offline correspondence proposed in \cite{Nakkiran:2020}, which focuses on datasets without noise. This study empirically showed a correspondence between online optimization and offline generalization for modern deep NNs trained to classify images. Earlier studies have proposed a similar comparison for linear models focusing on the asymptotic regime of training (see, e.g, \cite{bottou-lecun-2004,bottou-lecun-2004a}).

Contrastive learning, introduced by \cite{Chopa2005,Hadsell2006,oord2018representation}, has become one of the most prominent supervised \cite{khosla2020supervised, gunel2020supervised} and self-supervised \cite{bachman2019learning, tian2020contrastive, he2020momentum, chen2020simple} ML techniques to learn similarity relations of high-dimensional data, producing impressive results in several fields, see, e.g. \cite{le2020contrastive, jaiswal2021survey}. Despite its success, \cite{ohri2021review,liu2021self,le2020contrastive} show that contrastive learning usually requires huge datasets and considerable use of data augmentation techniques. Dealing with augmentation techniques and unlabeled data where negative samples are randomly selected introduces instance discrimination challenges, i.e., the need to find ways to limit the appearance of faulty positive and negative samples. Indeed, \cite{robinson2021can,wang2021understanding} show that the contrastive loss does not always sufficiently guide which features are extracted.
For these reasons, several works tackled the problem of discriminating against faulty negatives, as \cite{huynh2022boosting,kalantidis2020hard,chuang2020debiased}%,iscen2018mining}
, removing faulty positives and negatives dynamically (see \cite{robinson2021can,zhu2021improving}) and creating more robust contrastive setups introducing new losses (see \cite{chuang2022robust,morgado2021robust}) or architectural components, \cite{grill2020bootstrap}.

\section{Dataset construction}
\label{sec:contrastive}

In this section, we describe the choices we made to study the dataset features that influence training and generalization in similarity learning, i.e., the density of the image pairs and the presence of noise.
We start by defining the criteria  used to construct the pairs. 

\begin{figure*}[!h]
\center
%\vspace{-0.4cm}
%\vskip 0.1in
\begin{minipage}{\linewidth}
        \begin{subfigure}[b]{\textwidth}
        %\caption{Caption a}
        \begin{center}
        \begin{small}
        \begin{sc}
        \begin{tabular}{ m{3.1cm}  >{\centering\arraybackslash} m{3.5cm}  >{\centering\arraybackslash}m{3.5cm}   >{\centering\arraybackslash}m{3.5cm}}
        \toprule
        ~        &       No Noise       & PLN       &           SLN     \\
        \midrule
        &&&\\[-10pt]
        Dense Connections  & \includegraphics[height=0.12\textwidth]{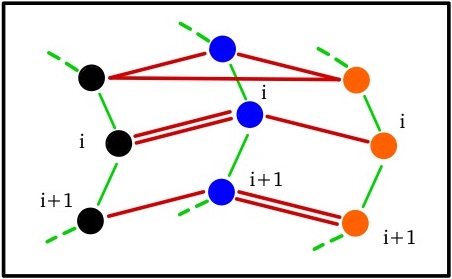}  & \includegraphics[height=0.12\textwidth]{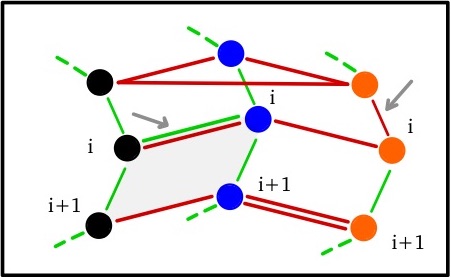} & \includegraphics[height=0.12\textwidth]{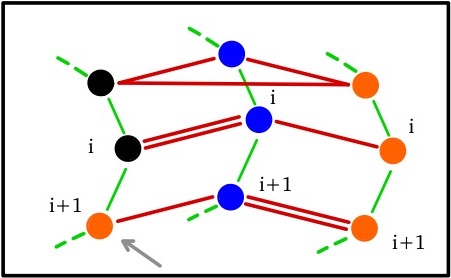}  \\
        Sparse connections & \includegraphics[height=0.12\textwidth]{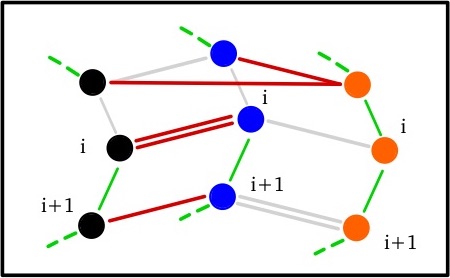} & \includegraphics[height=0.12\textwidth]{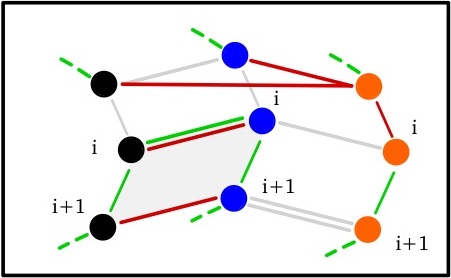} & \includegraphics[height=0.12\textwidth]{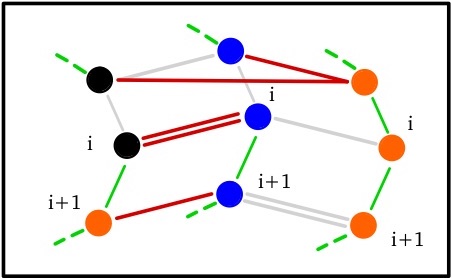}\\
        \bottomrule
        &&&\\[5pt]
        \end{tabular}
        \end{sc}
        \end{small}
        \end{center}
        \end{subfigure}
\end{minipage}
\vspace{-15pt}
\caption{Pictorial view of data relation appearing in Scenario 2 (top) and 1 (bottom) for different classes of data (black, blue, and orange). Each image corresponds to a node in the similarity graph whose color is representative of its class. Positive (similar) pairs are connected by green edges, negative (different) pairs by red edges. Ignored connections are in light gray. Multiple edges between two nodes refer to repeated pairs. Gray arrows indicate where noise appears, shaded areas (PLN column) are examples of transitivity breaking (DIBS). As discussed in Sec.\,\ref{sec:contrastive}, we create closed chains of positive pairs within the same class $c$, while the negative pairs are formed by connecting each image in $c$ to the element with the same index $i$ in a different class $c'$.\label{fig:similarity_graph} }
\end{figure*}

\vspace{-8pt}
\paragraph{Similarity graph.}
As opposed to classification problems, where the main concerns during dataset creation are class balancing and image diversity, in contrastive learning, we should consider that pair (or group) relations between images define an unoriented similarity graph inside the input space.
Calling $N$ the total number of images in the full dataset and  $N_{\rm pairs}$ the number of pairs, the density of this graph, $\rho=N_{\rm pairs}/\binom{N}{2}$, tells us the extent of knowledge we have about the input images. To enhance our understanding of a given dataset, we ought to create all possible labeled pairs, $\binom{N}{2}\sim N^2$, but this quickly becomes unfeasible when considering large datasets. For this reason, we construct pairs in a way that maximizes the information about similar images (all similar images are transitivity-related) and scales linearly with $N$. In practice, we construct closed chains of positive pairs within the same class, $c$, $\{ \{x_1^{c},x_2^{c} \}, \ldots, \{x_k^{c},x_{k+1}^{c}\},  \ldots, \{x_{n}^{c},x_{1}^{c}\}\}$, where $n$ is the total number of images in $c$. Then, to build negative pairs, each image is connected to the element having the same index in a different class, chosen at random $\{x_k^{c},x_{k}^{c'}\}$ where $c'\neq c$. If the original dataset classes are balanced, each image appears on average in 4 different pairs (2 times in the positive and 2 times in the negative pairs). Therefore, the total number of pairs is given by $N_{\rm pairs} = 2\times N=  2\times N_c \, \times \,n_c\,$, where $n_c$ is the total number of classes, and $N_c$ is the number of elements per class. Finally, we describe the dataset construction method we used to study how density in the similarity graph affects training.
\vspace{-0.2cm}
\begin{itemize}
    \item \textbf{Scenario 1: sparse connections.}  To train the network in the absence of noise, we first create the pairs using the full dataset. We follow the procedure described at the beginning of this section so that $N_{\rm pairs} = 2\times N$. We then take $N_{\rm sample}$ balanced pairs (data used to train the model) from the $N_{\rm pairs}$ list to train the NN and repeat this procedure  $n_{s}$ times. 
    
    \item \noindent \textbf{Scenario 2: dense connections.}  In this setup, we create a reduced version of the original dataset. Being interested in training the network on $N_{\rm pairs}$ pairs, we select $N_{\rm reduced}=N_{\rm pairs}/2$ images from the original training set. The reduced dataset is balanced so that we have $N_{\rm pairs}/(2 n_c)$ images per class. Then, we create our training and test samples using the same prescription described at the beginning of this section. We connect adjacent images within the same class, and each of them with a random image with the same index in a different class.  This way, we get exactly $N_{\rm pairs}$ pairs that will be automatically balanced. We repeat this procedure $n_{s}$ times.
\end{itemize}

Pictorial views of the similarity graph are shown in Fig. \ref{fig:similarity_graph}, where we represent elements belonging to different classes with nodes of different colors (black, blue, and orange classes), and similarity and dissimilarity relations with green and red edges, respectively. We will show that the qualitative relation between generalization and dataset density is independent of the specific method used in pair construction. In short, the relevant quantity is the probability of finding closed paths in the similarity graph. However, the approach used in this work allows for dealing with the problem analytically, as shown in Sec. \ref{sec:DD}.
\vspace{-8pt}

\paragraph{Noise introduction.}
SNNs can be subjected to different types of noise having different properties. To show their impact on the training process, we introduce two simple representatives, namely Single Label Noise (SLN) and Pair Label Noise (PLN), which are described below (see Fig. \,\ref{fig:similarity_graph} for an illustration).
\begin{itemize}
    \item \textbf{Single Label Noise (SLN).} Let us consider a dataset with $N$ samples   belonging to   $n_c$ classes and their corresponding labels $Y^S= \{y_1^S, y_2^S, \ldots , y_N^S \}$. Suppose the classes are uniformly populated $N_c=N/n_c$. If some label noise is present in the original dataset, this will propagate to the training pairs as these are created. If SLN is uniformly introduced across all classes, it will keep the original class balancing on average (over multiple samples). On the other hand, in every single run, statistical fluctuations of uniform distribution introduce some asymmetry in the original class representative number (see Fig. \ref{fig:similarity_graph}). Finally, in the presence of SLN, similarity relations (reflexive, symmetric, and transitive properties) are preserved as mislabeling appears in all pairs containing a misclassified image.
    
    \item \textbf{Pair Label Noise (PLN).} Let us now consider a dataset of $N_{\rm pairs}$ pairs  with pair labels  $Y^P= \{y_1^P, y_2^P, \ldots , y_{N_{\rm pairs}}^P \}$, which can be similar ($y^P=1$) or different ($y^P=0$). We construct them so that they are balanced (half are similar, and half are different). Suppose we randomly shuffle some fraction of the total labels. In that case, the noise we introduce is symmetric under similar $\leftrightarrow$ different changes, and it acts democratically on every class of the original dataset. On the other hand, PLN  can lead to inconsistent relations in the pairs dataset. Indeed, as we will show in the following sections, it breaks transitivity and, therefore, similarity. 
\end{itemize}

As discussed later, these two sources of noise impact  how models learn similarity relations in distinct ways. To fairly compare the outcome of the model in the presence of PLN and SLN, we need to ensure that we introduce the same amount of input label noise in the two setups. We present below how we ensured that the same amount of \emph{effective noise} was introduced. Being $n_c$ the number of image classes, $y_i^S$  the label of the $i$-th image, and $y^P_i$ the label of the $i$-th pair of images, we can define the SLN transformation as
\be
\label{eq:SLN}
\mc{T}_{\text{SLN}}(q):y_i^S\rightarrow \mbox{random}(1,n_c) \quad\mbox{with probability}\;q
\ee
and the PLN transformation as
\be \label{eq:PLN}
\mc{T}_{\text{PLN}}(\tilde q):y^P_i\rightarrow \mbox{random}(0,1)\quad\mbox{with probability}\;\tilde{q}\,.
\ee
As SLN appears in the dataset before pair creation and the pairs are constructed so that the dataset is balanced (half pairs are similar, half are different), the probability of effective pair mislabeling induced by SLN, $P_{\textrm{SLN}}(q)$, is given by
\be
P_{\rm{SLN}} (q)= q-\frac{q^2}{2}\,.
\ee
while the probability of effective pair mislabeling coming from PLN, $P_{\rm{PLN}}(\tilde{q})$, is
\be
P_{\rm{PLN}} (\tilde q)=\frac{\tilde q}{2}\,.
\ee
The requirement of having the same amount of effective noise in the dataset ($P_{\rm{SLN}} (q)= P_{\rm{PLN}} (\tilde q)$) boils down to the following relation between $q$ and $\tilde q$: \be
\label{eq:samenoise}
q= 1 - \sqrt{1-\tilde q}.
\ee
The details of this derivation and the pseudocodes describing dataset creation can be found in the Supplementary Material \ref{app:prob} and \ref{app:create_pairs}. 
\vspace{-0.2cm}
\subsection{Experimental setup}
\label{sec:experiments}
\begin{figure*}[!ht]
\center
 \begin{sc}
    \includegraphics[height=0.14\textwidth]{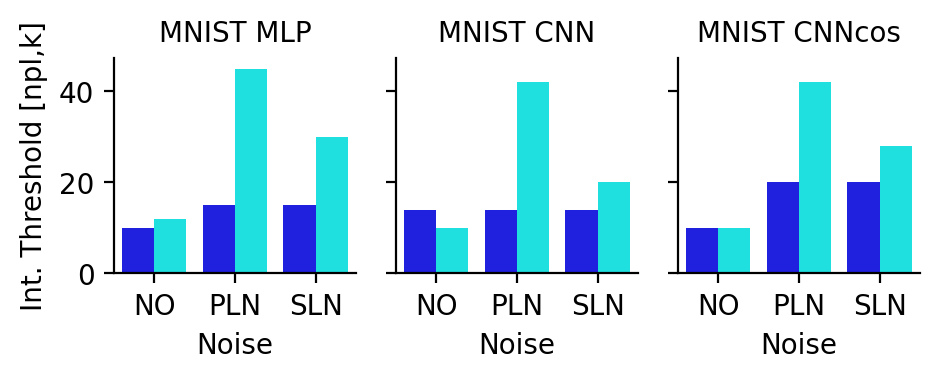} \includegraphics[height=0.14\textwidth]{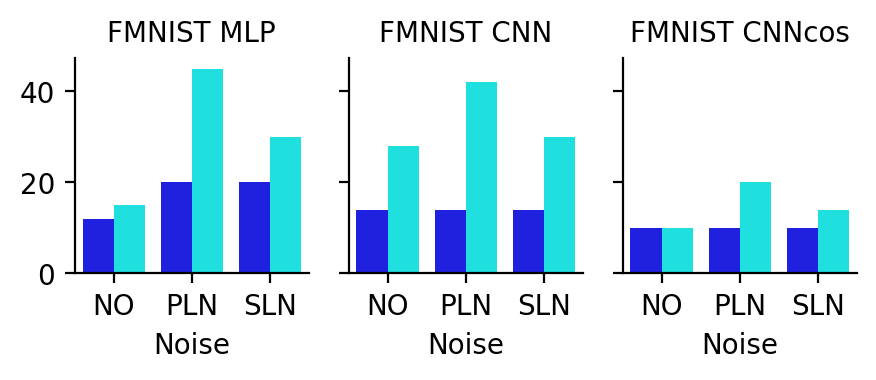} 
    \includegraphics[height=0.14\textwidth]{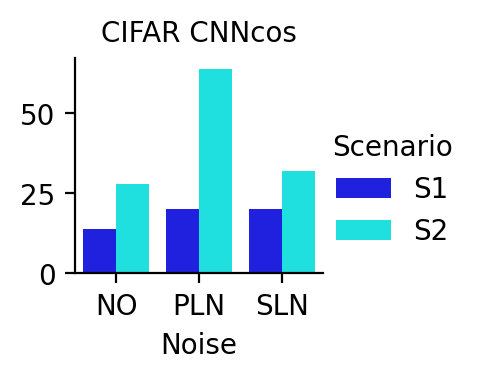} \\
    \includegraphics[height=0.14\textwidth]{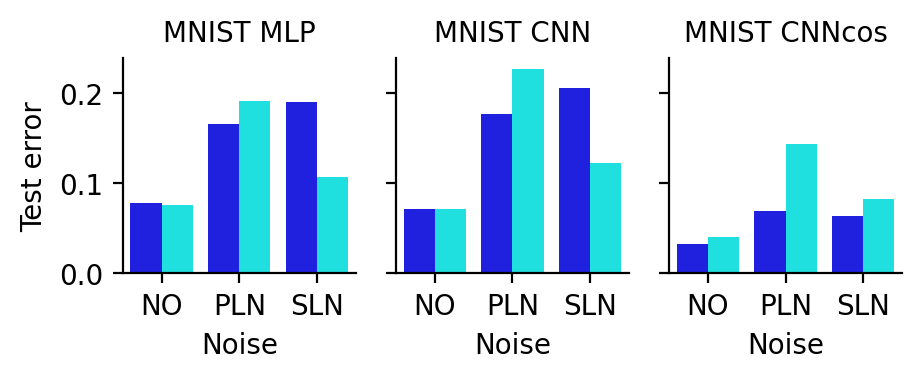} \includegraphics[height=0.14\textwidth]{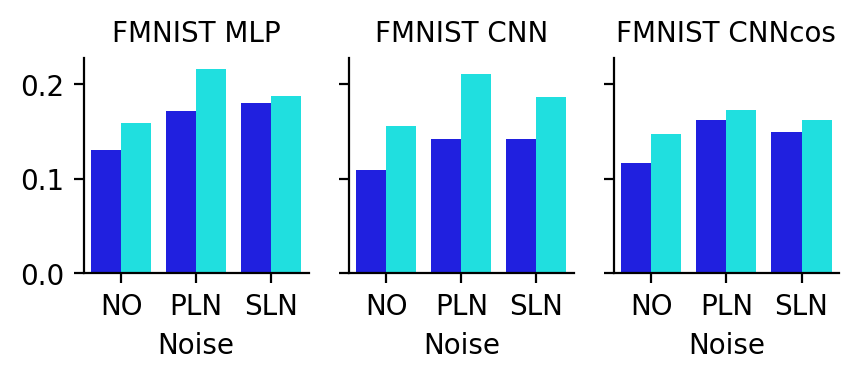} 
    \includegraphics[height=0.14
\textwidth]{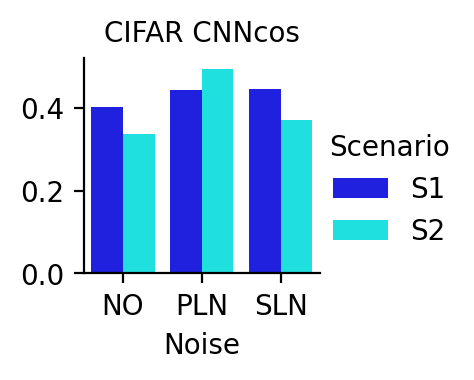}
    
\end{sc}

  \caption{\textbf{Top:} Relation between training setup (see Sec.\,\ref{sec:experiments}), noise source (`NO' refers to the scenario without noise), and interpolation threshold (DD peak location) expressed using number of neurons per layer (npl) for MLP setups or number of filters, $k$, for CNN cases. Note that PLN in S2 typically requires more parameters to interpolate. \textbf{Bottom:} Average test errors in the deep overparametrized regime  after 2000 epochs. Due to the breaking of transitivity (DIBS phenomenon), PLN  average test errors in S2 stay higher in the deep overparametrized regime.  \label{fig:histograms}  }
\end{figure*}

%architectures
In this work, we consider two Siamese branch architectures. The first one is an MLP with 3 hidden layers having the same width and ReLU activation functions with Xavier uniform initialization, see \cite{glorot2010understanding}. The second architecture is a 4-layer CNN. We also considered two training setups: in one case, we compute the Euclidean distance in the output layer training the network using Contrastive Loss from \cite{Hadsell2006}, and in the other one, we compute the cosine similarity training the network using Cosine Embedding Loss (see Supplementary Material \ref{app:distances_and_losses} for details). The CNN architecture is based on the model described in \cite{Page2018}, it contains three Convolution-BatchNormalization-ReLU-MaxPooling layers and a fully-connected output layer. The number of filters in each convolution layer scales as [$k$, 2$k$, 2$k$] while the MaxPooling is [1, 2, 8]. We fix the kernel size = 3, stride = 1 and padding = 1. When we train the network using contrastive loss (cosine embedding loss), we set the fully-connected output layer width to $k$ (2$k$).

\vspace{+0.1cm}
\textbf{Double Descent (DD) setup.}
We test the presence of DD using MNIST \cite{lecun2010mnist}, FMNIST  \cite{xiao2017/online} and CIFAR10 \cite{cifar10} datasets. To understand the impact of overparameterization, we study how training and test errors vary at increasing network width and training time. To do so, we increase the number of neurons per layer in the fully connected architecture and the parameter $k$ in the CNN.  For all datasets, we consider 6000 training and 9000 test pairs. In every DD experiment, we let the network evolve for 2000 epochs using Adam optimizer with minibatches of size 128 and learning rate $\lambda=10^{-4}$, except when explicitly stated. All the hyper-parameters and the margins were chosen empirically. To see the average effect regardless of the particular choice of images in the dataset and weights initialization, we run 15 evolutions of the network using different training and test samples at each time. In most of the experiments, unless otherwise stated, we considered $\tilde q=0.2$, i.e., an effective noise of $10 \%$.

\vspace{+0.1cm}
\textbf{Online/offline setup.}
Since we cannot reuse samples for the online training, we consider an extended version of the standard MNIST dataset, namely the EMNIST (from \cite{emnist}). We use the digit section of EMNIST  that contains 240,000 training (and 40,000 test)  28 $\times$ 28  greyscale pixel images.  We train the offline case (Real world) over 40 epochs using 12k pairs that are created considering  sparse and dense scenarios. The online scenario (Ideal world) is trained once on 480k pairs created using the full training set of 240k samples. We test the models with 9k pairs constructed from the test set and consider Siamese networks with MLP and CNN blocks described at the beginning of this section. In order to compare the results on different network architectures, we used a comparable total number of parameters, namely, 200 nodes per layer for the MLP cases (total of 237,400 parameters) trained with the contrastive loss ($\lambda= 10^{-4}$); and width $k=47$ (total of 235,611 parameters) for the CNN cases trained with the cosine loss ($\lambda$  $ = 5\times 10^{-5}$). To provide an estimate of the results regardless of the particular choice of images and network initialization, we run MLP (CNN) experiments 5 (4) times. 
%All our experiments make use of the TensorFlow/Keras framework \cite{tensorflow2015-whitepaper}.
Each of the experiments mentioned above was performed in the presence and absence of noise and considering sparse (scenario 1) and dense pairs (scenario 2) in the training set. 

\section{Results}
\label{sec:DD}
\textbf{Double Descent (DD) results.} In all experiments we see the DD behavior, regardless of architecture, loss function, scenario and noise.  This does not happen in classification problems which typically require the presence of noise to make the DD curves clearly visible (see, e.g., \cite{NakkiranDeepDD:2020}).
As expected, DD becomes more prominent in the presence of noise. At the bottom of Fig.\,\ref{fig:PLN_SLN}, we show how the network reacts to different amounts of noisy labels.
In \textbf{Scenario 1}, the input dataset connections are sparse, and PLN and SLN have the same impact on training. This is understandable as there should not be any difference between PLN and SLN effects in the extreme case where every image appears only once in the training set. 
Instead, \textbf{Scenario 2} is characterized by dense input connections, and the system behaves differently under SLN and PLN. We experimentally observe that the DD peak location changes between PLN and SLN in almost all setups considered, see the rightmost plot at the bottom of Fig.\,\ref{fig:PLN_SLN}  and the top of Fig.\,\ref{fig:histograms}. Specifically, PLN peaks are shifted to the right-hand side, hinting that PLN  is harder to interpolate than SLN as it requires more parameters.  Increasing the amount of noise enhances the test errors as expected, but does not induce any significant peak shift. 

SLN test error tends to be higher in small to medium network sizes, see Fig.\,\ref{fig:PLN_SLN}. A hint about how this happens is given in Fig.\,\ref{fig:similarity_graph}. Indeed, SLN introduces a systematic error: a mislabeled image appears to be mislabeled in every pair. Therefore, given that the image features are not going to agree with pair labels, the only way the network has to classify correctly is by extracting the image from its natural distribution. NNs being continuous functions, this implies that a neighborhood of said image must be extracted as well, increasing the test error. At higher network widths, the volume of the mislabeled image neighborhood can become arbitrarily small, and the test error is free to go down again. In fact, \emph{SLN introduces systematic errors that do not compromise the consistency of the similarity graph}. On the other hand, PLN stays higher in the deep overparametrized regime (see bottom plots in Fig. \ref{fig:histograms}). Indeed, randomly changing similarity relations in the input dataset, \emph{PLN ends up breaking transitivity, making the training set similarity graph inconsistent}. Beyond keeping test error high, this inconsistency also implies that the network is never able to overfit completely: the training error will no longer vanish just by increasing the number of network parameters, see e.g., train error plots at the bottom of Fig.\,\ref{fig:PLN_SLN}. This effect is exacerbated when using dense datasets.

\begin{figure}[!ht]
\center
    \includegraphics[height=0.35\linewidth]{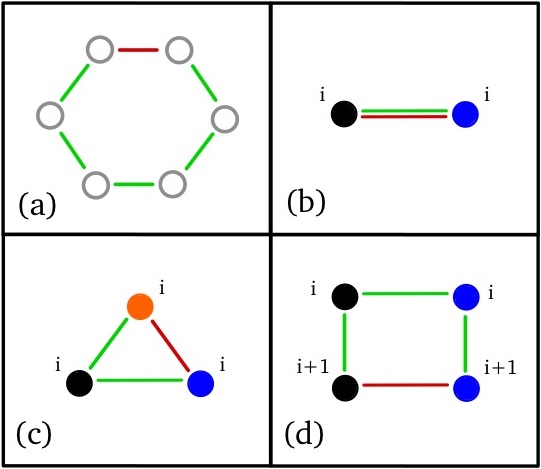}
  \caption{Similarity breaking configurations (a) and their leading contributions (b,c,d). \label{fig:DIBS_figs}  }
 % \vspace{-0.1cm}
\end{figure}

\begin{figure*}[!ht]
\center
 \begin{sc}
    \includegraphics[height=0.21\linewidth]{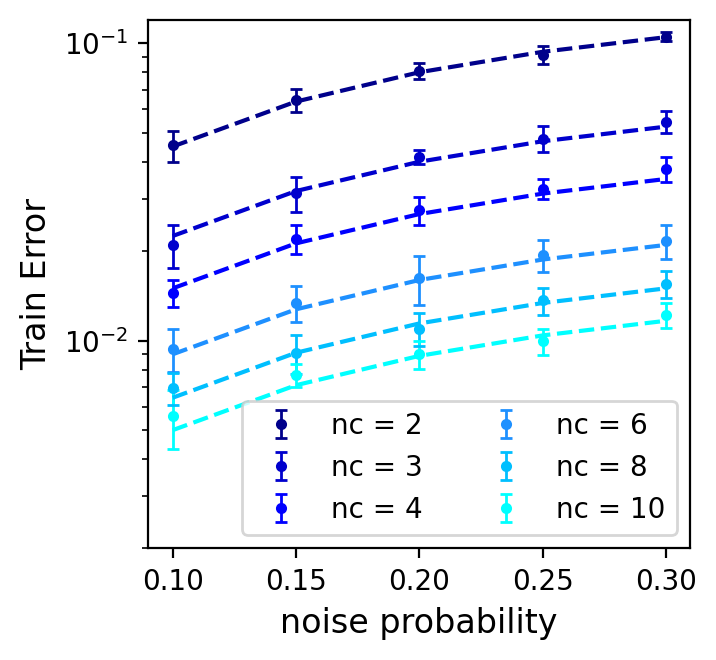}
    \includegraphics[height=0.21\linewidth]{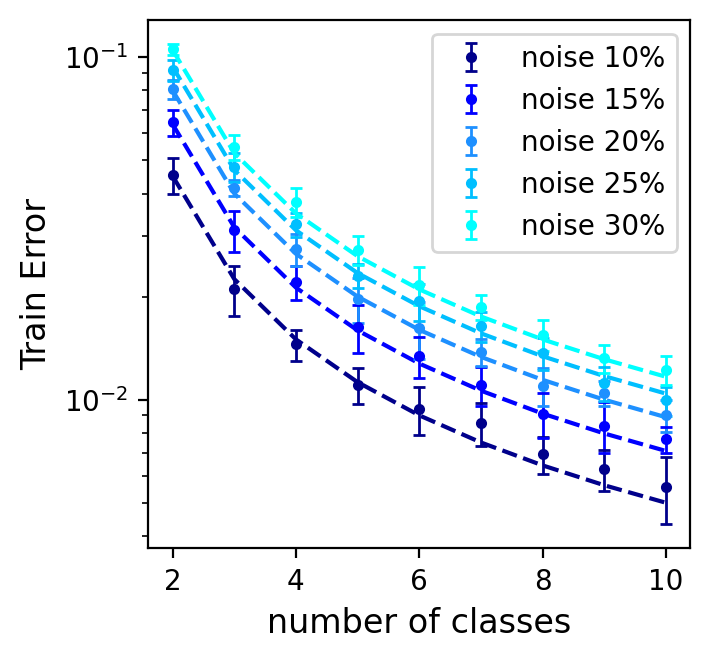}
    \includegraphics[height=0.21\linewidth]{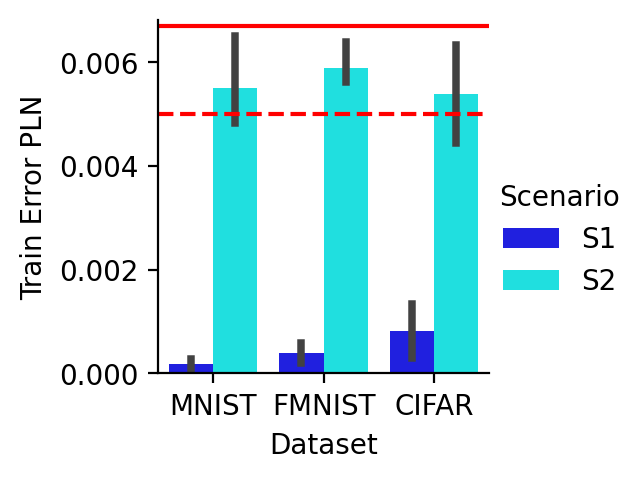}
\end{sc}
 % \vspace{-10pt}
  \caption{Analytic 1st order (dashed lines) and numerical (scatter points) estimates of the asymptotic training error behavior at varying number of classes $n_c$ (\textbf{left}) and effective noise (\textbf{center}) in the presence of PLN in Scenario 2 (Dense) for the FMNIST dataset trained on the MLP architecture with 500 neurons per layer, using Euclidean distance and contrastive loss. \textbf{Right}: Comparison between experimental training error distributions and lower (dashed line) and upper (solid line) bounds of Theorem \ref{th:first}. The DIBS phenomenon is observed in all datasets considered. \label{fig:DIBS}  }
 % \vspace{-0.1cm}
\end{figure*}
\begin{figure*}[!ht]
\center
 \begin{sc}
    \includegraphics[width=0.33\textwidth]{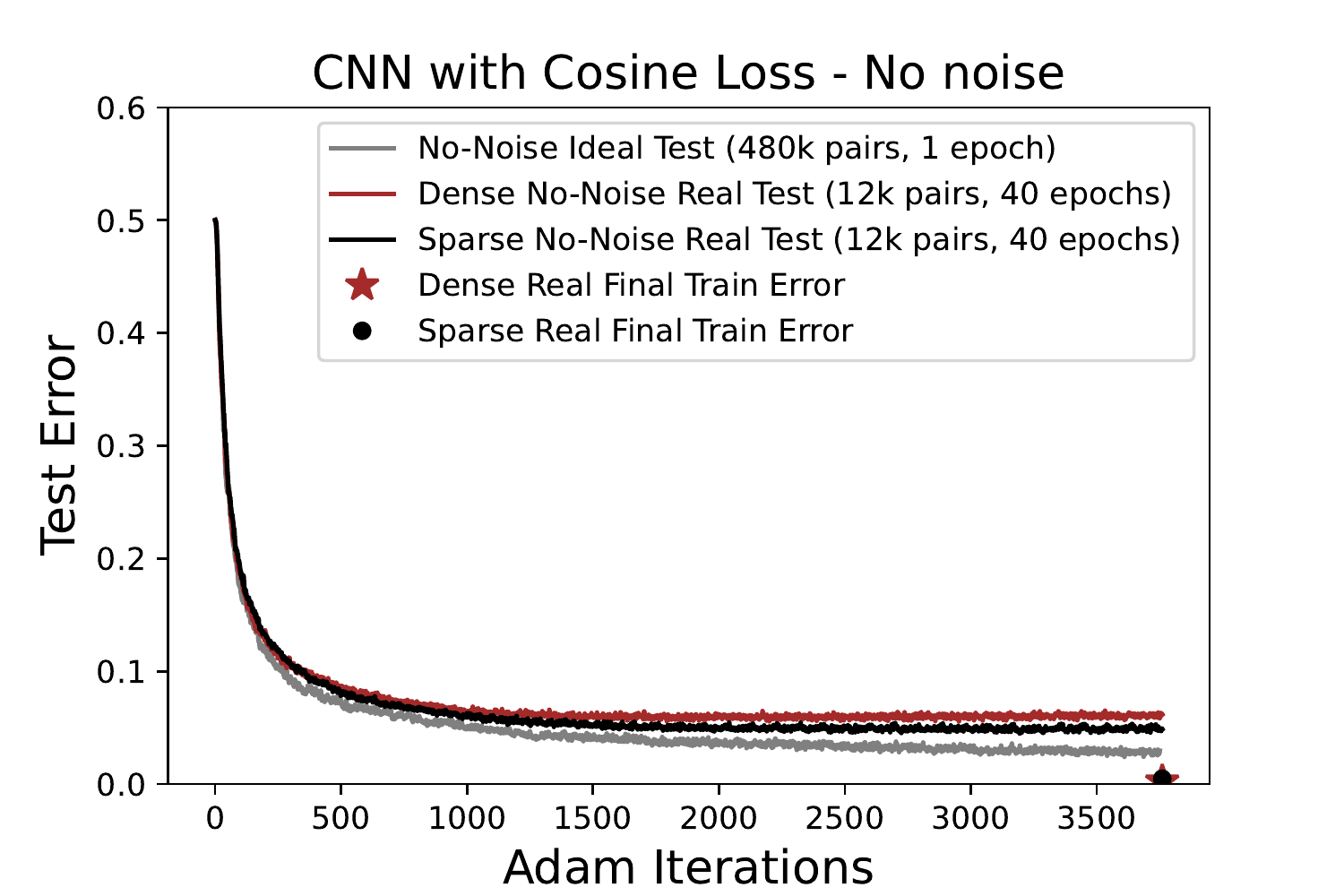}
  \includegraphics[width=0.33\textwidth]{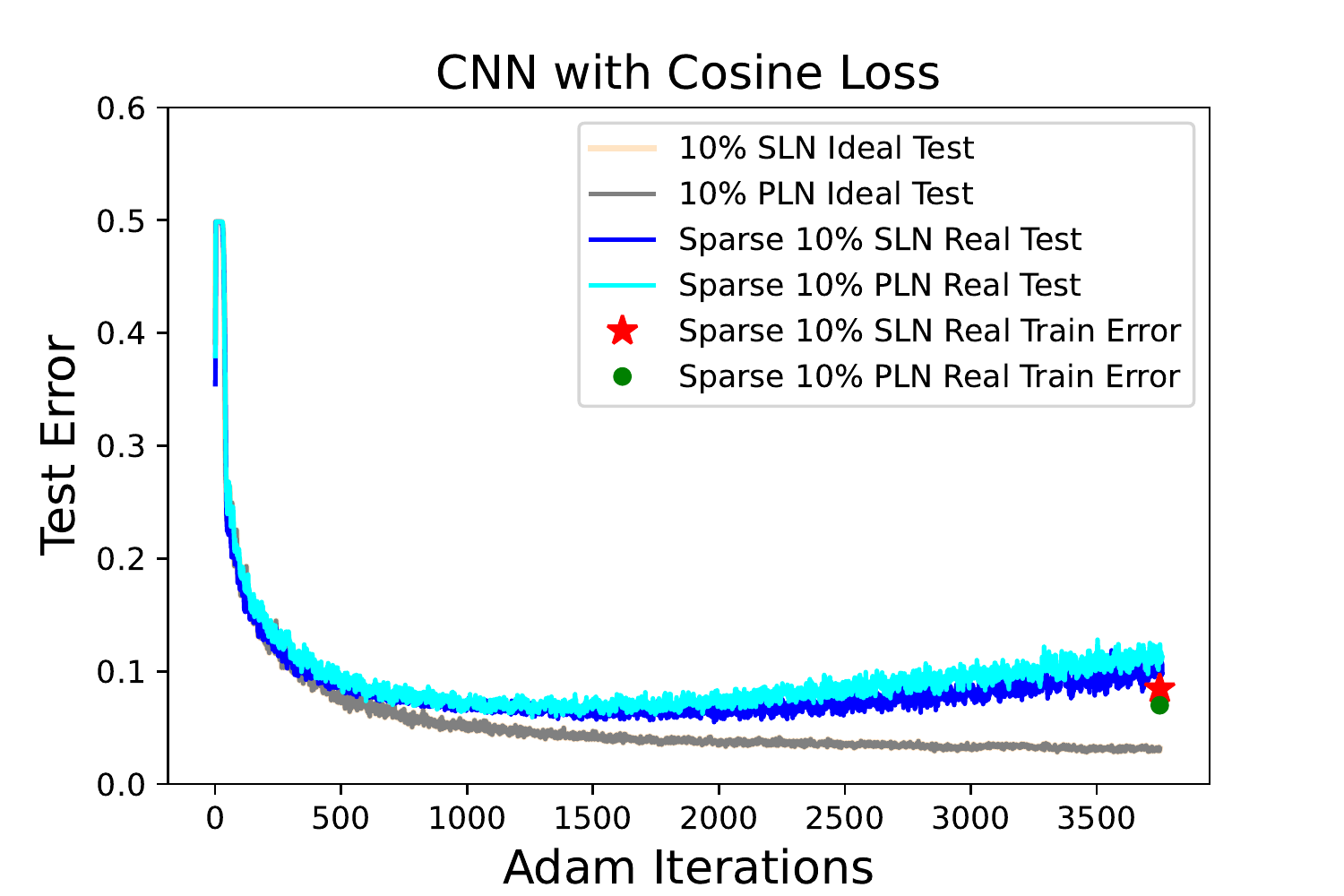}
\includegraphics[width=0.33\textwidth]{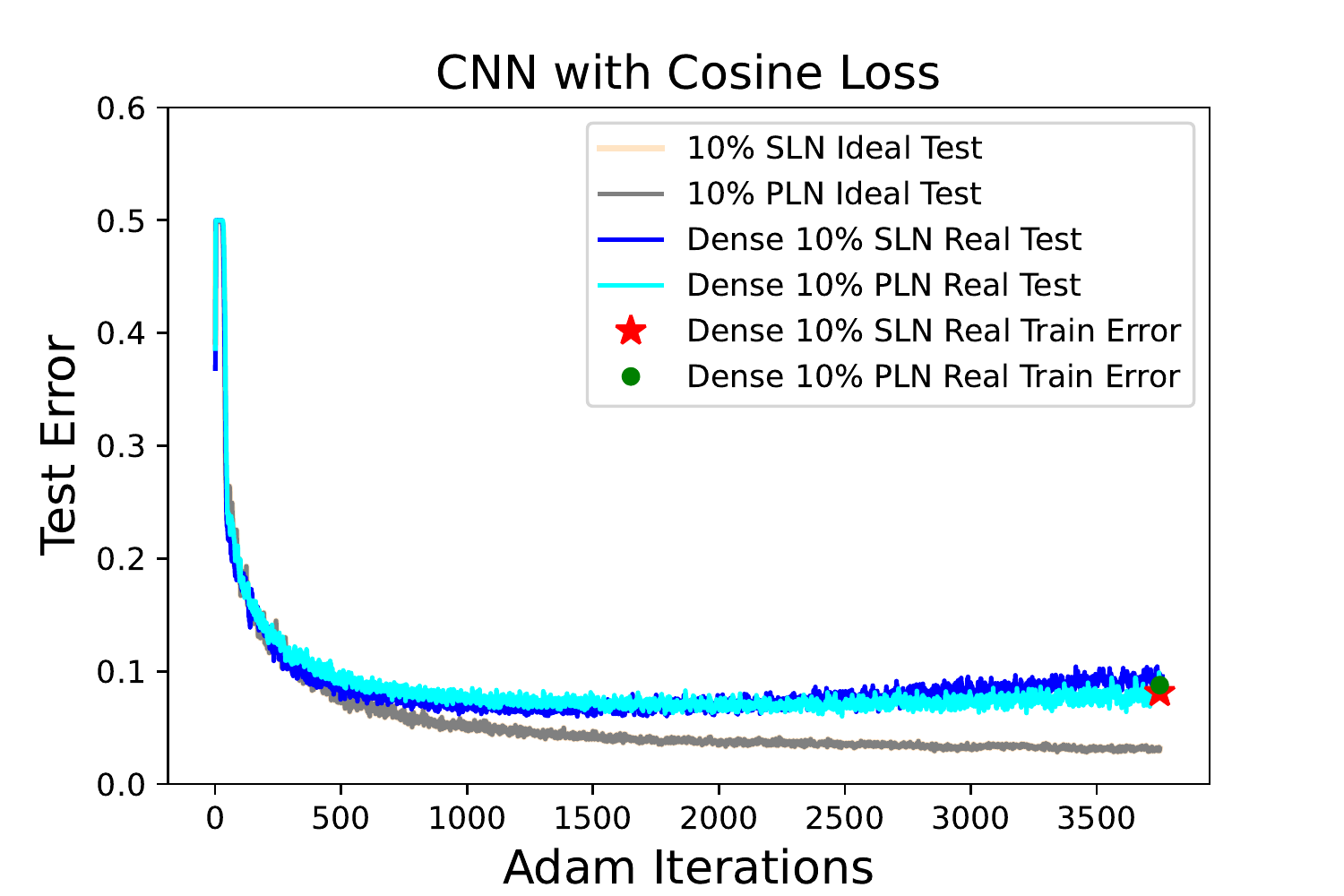}
\end{sc}

  \caption{Ideal (online) \emph{vs.} sparse/dense Real (offline) worlds for the  CNN architecture  with $k=47$ trained with the cosine loss in the absence of noise (\textbf{left}) and with 10\% of label noise for Scenario 1 (Sparse) (\textbf{center}) and Scenario 2 (Dense) (\textbf{right}).  Plots show the  median test
errors as a function of minibatch Adam iterations.  The stars (dots) correspond to the median real-world training errors at the end of training. \label{fig:bootstrap} }
\end{figure*}
\vspace{-9pt}
\textbf{Origin and magnitude of DIBS.}
We now explain the origin of the phenomenon we call \emph{Density-Induced Break of Similarity} originating from PLN. We can see if similarity is satisfied or violated in the training set by evaluating the consistency of the closed paths in the training pairs graph. Examples of inconsistent paths are depicted in Fig.\,\ref{fig:DIBS_figs}. Similarity breaking on 2-paths, configuration (b), corresponds to symmetry breaking, while on $n-$paths, as in configurations (a,c,d), where $n>2$,  corresponds to transitivity breaking. 
\vspace{-0.4cm}
\begin{theorem}
\label{th:first}
Let $\mathcal{D}$ be a dataset containing elements belonging to $n_c$ classes, each having $N_c$ elements. Let $x_i^c$ denote the $i$-th element of the $c$-th class. Let $\mathcal{G}_S$ be the similarity graph induced by the creation of similar and dissimilar data pairs. Let the positive pairs be constructed as $\{x_i^c,x_{i+1}^c\}$, so as to form closed chains of similar pairs. Let the negative pairs be constructed as $\{x_i^c,x_{i}^{c'}\}$, $c'\neq c$, so as to generate random graphs of dissimilar pairs between elements of the same index. If the transformation $\mathcal{T}_{\rm{PLN}}(2P)$ is applied to the pairs thus created, it induces the break of similarity resulting in an asymptotic training error, 
\be
{\rm{Error}}_{\rm{DIBS}}=\lim_{n_{\theta}\rightarrow \infty} {\rm{TrainError}}_{\rm{{\footnotesize{Dense}}}}^{\rm{\footnotesize{PLN}}} (P,n_c)\,,
\ee 
that is limited by:
\begin{equation}
\frac{P\,(1-P)}{2(n_c -1)}\leq \rm{Error}_{\rm{DIBS}} - \mathcal{E}_{sim} < \mathcal{E}_{diff}
\label{eq:asymptotic_tr_err}
\end{equation}
where $n_{\theta}$ is the number of network parameters, $P$ is the probability of effective pair mislabeling induced by PLN,  $\mathcal{T}_{\rm{PLN}}$ is the transformation given in Eq.\,\ref{eq:PLN}, and 
\begin{equation}
\begin{array}{lll}
\mathcal{E}_{{\rm sim}} = &  \displaystyle\frac{P(1-P)^{N_c-1}}{2}\,,\\[10pt]
\mathcal{E}_{{\rm diff}} = & \displaystyle\sum_{m=2}^{nc} \frac{m\,P^{m-1}(1-P)}{2^{m}(n_c-1)^{m-1}}\frac{(n_c-2)!}{(n_c-m)!}\\[10pt]
& \displaystyle\times \sum_{i=0}^{N_c/2} \left[\frac{(1-P)}{2}\right]^{2i}.
\vspace{0.1cm}
\label{eq:EeqEdiffEmix}
\end{array}
\end{equation}
\end{theorem}
\begin{corollary}
If the assumptions of Theorem 1 hold and $n_c=2$ then:
${\rm{Error}}_{\rm{DIBS}} - \mathcal{E}_{{\rm sim}}  = \frac{P\,(1-P)}{2}$.
\label{cor:frist}
\end{corollary}
\vspace{-0.1cm}
Below we briefly list the highlights of the proof of the theorem (and the corollary) to be analyzed to derive the above results.
The upper bound in Theorem \ref{th:first} can be proved by noticing that reflexivity and transitivity breaking can only appear in:
\vspace{-0.1cm}
\begin{enumerate}
\item Closed chains of similar pairs containing only 1 mislabeled pair (as in Fig.\,\ref{fig:DIBS_figs}, in configuration (a), when all nodes belong to the same class);
\item Random closed $n$-paths containing elements of different classes with the same index (as in configurations (b,c));
\item $n$-paths containing multiple classes and more elements of the same class (as in configurations (d)).
\vspace{-0.1cm}
\end{enumerate}
The lower bound in Theorem \ref{th:first} is the first-order approximation of the upper bound coming from 2-path only. The inequality can be proved by showing that there is no one-to-one correspondence between inconsistent $n$-paths and the number of classification errors. More graph inconsistencies can lead to a single error. To facilitate the counting of the unavoidable errors associated with a given configuration, one can resort to collapsed configurations, i.e., collapsing nodes connected by green vertices, and then counting the number of inconsistent 2-paths. In particular, the Corollary \ref{cor:frist} follows from noticing that, if $n_c=2$, the similarity-breaking contribution coming from point 3) is completely redundant with that of point 2). The formula associated with the lower bound, and thus to inconsistent 2-paths, is given by the probability of having two elements, with the same index and belonging to different classes, connected by a correctly classified different pair and a noisy one. This leads to:
\[
 \underbrace{\frac{N_{\rm diff. pairs}}{N_{\rm pairs}}\times P}_\text{noisy different pair}  \overbrace{\frac{N_{\rm diff. pairs}}{N_{\rm pairs}}\times (1-P)}^\text{correct different pair} \underbrace{\frac{\overbrace{2}^\text{\# configurations}}{n_c - 1}}_\text{connected to same 2 classes}\]

\vspace{0.1cm} 
\noindent where $N_{\rm diff. pairs}$ is the number of different pairs, and $N_{\rm pairs}$ is the total number of pairs, with $\frac{N_{\rm diff. pairs}}{N_{\rm pairs}}= \frac{1}{2}$ as we consider balanced equal/different pairs. Finally, it is easy to see that the contribution coming from  $\mathcal{E}_{{\rm sim}}$ is negligible in standard situations where $N_c \gg 1$.
In Fig. \ref{fig:DIBS}, we validate our formula by comparing it with experimental results. In particular, in the left and central panels,  we consider the FMNIST dataset trained on our MLP architecture with 500 neurons per layer, using Euclidean distance and contrastive loss (see Sec.\,\ref{sec:experiments}). Numerical results (mean and standard error bar) come from 10 runs where we choose different random classes each time.  These results show that, in the overparametrized regime, the training error follows the behaviors of solid lines given by the lower bound of Theorem \ref{th:first}. The right panel of Fig.\,\ref{fig:DIBS} shows the mean and standard deviation of the training errors in the overparametrized regime obtained in all our DD experiments in scenarios 1 and 2. Moreover, we compare them with the lower (red dashed line) and upper bound (solid red line) of Theorem \ref{th:first} for 10 classes and $P=0.1$. We find perfect agreement between experiments and theoretical results. 
% \vspace{-0.1cm}

This analysis shows that the macroscopic presence of transitivity breaking is linked to the presence and number of closed paths in the similarity graph and therefore to the dataset density. 

\vspace{0.1cm}
\textbf{Online (Ideal world) \emph{vs.} Offline (Real world) learning.}
We probe the correspondence between offline generalization and online optimization \cite{Nakkiran:2020} for similarity tasks by studying how the training setting and the presence of noisy labels can impact these two regimes. Considering usual training settings (i.e., natural choices of architecture-loss function match), the conjecture holds for data without noise, regardless of the dataset density. See the left panel of Fig.\,\ref{fig:bootstrap} for the CNN architecture equipped with cosine loss in the absence of label noise (experimental details were given in Sec.\,\ref{sec:experiments}). In the presence of noise, however, we find that the online/offline correspondence breaks down for all choices of training settings considered.

Two representative examples where the conjecture breaks are depicted in Fig.\,\ref{fig:bootstrap}. There, we show the median test error values on dense and sparse datasets of real- and ideal-world scenarios with 10\% of PLN and SLN trained using the CNN architecture. We compare offline and online settings after the same number of training iterations. We observe that while both ideal and real test errors are affected by noise,  this effect is exacerbated in the real world scenarios. In fact, we observe that the introduction of ``fresh'' samples to the model, even if they possess noisy labels, enhances the model's diversity and ultimately improves its generalization. Note that the ideal
world curves (gray and bisque) overlap with each other. Interestingly, we also find that the online/offline correspondence for similarity tasks is influenced by the network architecture and the loss function choice.  Nevertheless, independently of the architecture-loss matching, the equivalence between online and offline settings breaks down in the presence of label noise for all the scenarios considered.

\vspace{0.1cm}
\textbf{DIBS and modern contrastive learning.}
The similarity-breaking nature of PLN in dense datasets should not be underestimated as it may appear in widely employed settings. Modern approaches to self-supervised contrastive learning (see the recent reviews of \cite{ohri2021review,liu2021self,jaiswal2021survey,le2020contrastive}) heavily rely on data augmentation to learn representations \cite{tian2020contrastive}. The massive use of data augmentation, however, may result in partial representation learning (feature suppression) or lead to semantic errors as in \cite{purushwalkam2020demystifying}. Moreover, as exposed in \cite{huynh2022boosting}, if negative pairs are formed by sampling views from different images, regardless of their semantic information, this may lead to the appearance of false-negative pairs, potentially breaking transitivity and compromising the training efficiency. 
Interestingly, this skewness towards false-negative pairs is the same effect we find studying the asymptotic training error balance with DIBS. 
Notwithstanding these issues, data augmentation and random selection of negative samples are intrinsic to self-supervised methods.\footnote{For example, in a pretext task, the original image acts as an anchor, its augmentations act as positive samples, and the rest of the images in the batch (or in the training data) act as negative samples.} Therefore, several works in contrastive learning have focused on controlling the quality of augmented data and mitigating the effects of false negatives (see Sec.\,\ref{sec:related}). When two different images belonging to the same class  (sharing semantic features) are classified as different, convergence slows down and semantic information gets lost. This goes under the name of instance discrimination task (i.e., the problem of discriminating pairs of similar points from dissimilar ones), and failing it can harm the formation of features useful for downstream tasks. For this reason, feature extraction in self-supervised contrastive learning is usually affected by pair-label noise by construction.
\vspace{-0.2cm}
\section{Discussion and conclusions}
\label{sec:discussion}

We move the first steps towards understanding generalization in similarity learning focusing on SNNs. To do so, we borrow the frameworks of DD and online/offline correspondence from classification tasks. We show that DD appearance is magnified in SNNs as it appears also in the absence of noise. Notably, we find that noise and the density of pairs in the training set crucially affect generalization. We present two kinds of noise: SLN, preserving similarity relations, and PLN, breaking transitivity. The same noise sources presented in this work can be easily generalized to models where the network input is given by multiple images. Studying DD, we show that similarity-breaking noise compromises the asymptotic generalization performance (large training time) of the network in the overparametrized regime. Moreover, these effects get magnified at increasing training set density, preventing perfect interpolation. Studying the online/offline correspondence, we find that the generalization properties before overfitting time are not sensitive to the density of the training set and only depend on noise. In particular, in the presence of noise, the online/offline correspondence breaks down and the differences between the real and ideal generalization gap are not universal and depend on the training setup.

\vspace{0.1cm}
\textbf{Limitations.}
This is an exploratory work that does not investigate all possible setups which may affect or lead to DD, such as regularization (see \cite{nakkiran2020optimal,mei2022generalization}),  epoch and sample-wise DD (see, \cite{NakkiranDeepDD:2020,bodin2021model,heckel2020early,pezeshki2021multi}). Moreover, we focus on the under- and overparametrized regime without providing quantitative results about the interpolation threshold itself,  \cite{d2020double,d2020triple,mei2022generalization}. This is because, to the best of our knowledge, there is no predefined way of treating SNNs analytically as no proxy model as Random Fourier Features (see \cite{rahimi2007random}) can be constructed. Indeed, while in classification or regression tasks the output layer size is known, this is not true for SNNs. For this reason, we believe that an analytic study of DD in SNNs may require another approach, and we leave this study for future work.

\vspace{0.1cm}
\textbf{Outlook.} In the majority of modern contrastive learning works, the final graph of similarity relations in the dataset becomes really dense as each training step involves multiple images at a time. Moreover, from instance discrimination task examples, we know that contrastive learning tends to be affected by faulty positive and negative pair relations. This is the setting where we find that noise crucially impacts generalization. While the technological developments and the applications of contrastive learning kept expanding during the last years, a fundamental study about how it generalises and reacts to noise is still missing.

\ack NF acknowledges the UKRI support
through the Horizon Europe guarantee Marie Skłodowska-Curie grant (EP/X036820/1). This research was supported in part through the Maxwell computational resources operated at Deutsches Elektronen-Synchrotron DESY, Hamburg, Germany.
\vspace{-10pt}

\bibliography{ecai}

\appendix 
\onecolumn
\section{Supplementary Material}
\label{app:supp-material}

\vspace{0.3cm}
% \onecolumn %% Turn this off if single column is desired for the supplement
\subsection*{Table of contents}

\begin{itemize}

\item[] \textbf{\ref{app:create_pairs} Pseudocode to create the balanced pairs and reduced dataset}
\item[]  \textbf{\ref{app:prob}     Effective noise derivation}
\item[] \textbf{\ref{app:distances_and_losses} Output layer and loss function}
%\item \textbf{\ref{app:unbalanced_train_error} \st{Unbalanced train error with PLN DIBS} \nf{Kept it here until we remove it}}
%\item
%\textbf{\ref{app:online_offline_details} Online/Offline Correspondence}

\end{itemize}

\subsection{Pseudocode to create the balanced pairs and reduced dataset}
\label{app:create_pairs}
\begin{figure}[!h]
\center
  \includegraphics[width=\textwidth]{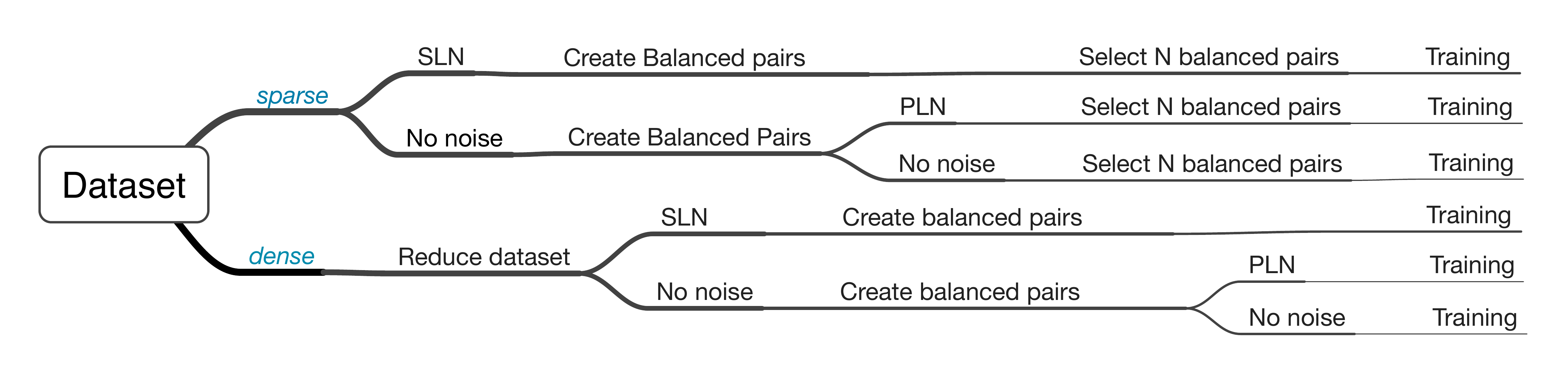}
  \caption{Pipeline guiding from the original dataset to training in the different setups considered. \label{fig:flowchart}  }
\end{figure}

Below, we introduce the pseudocode of the strategies used to introduce PLN and SLN  in Scenario 1 (sparse connections) and 2 (dense connections). In particular, Algorithm \ref{alg:create_pairs} explains how we create balanced positive and negative pairs, Algorithm \ref{alg:reduce_dataset} explains how we create a reduced and balanced version of a dataset. Algorithms \ref{alg:sparse_PLN} and \ref{alg:sparse_SLN} (\ref{alg:dense_PLN} and \ref{alg:dense_SLN}) describe the pipeline to train the network using sparse (dense) dataset relations in presence of PLN and SLN respectively.
A pictorial view of the paths leading to training in the
different setups considered is depicted in Fig. \ref{fig:flowchart}.

\begin{algorithm}[!h]
\caption{Creating Balanced  Pairs \label{alg:create_pairs}} 
\begin{algorithmic}
\Function{CreateBalancedPairs}{$X$, $Y_S$, $n_c$}
\State  pairs=[]
\State  $Y_P$=[]
\State  $M_d \gets [m_1,\dots m_{n_c}] $\hfill list of number of samples in each class
\State  $X_{N_d} \gets[X_1,\dots X_{n_c}] $ \hfill list of images in each class
\For{$d=1,2,\ldots,$ $n_c$}
\For{$i=1,2,\ldots, m_d$}
\State pairs\_pos = \{$X_d$[i], $X_d$[i+1]\}
\State $d_x$ =  ($d$ + random-integer(1,$n_c$-1)) \% $n_c $ \hfill select different random class 
\State $j$ =  random-integer(1,$m_{d_x}$) \hfill select random element in the class
\State pair\_neg =  \{$X_d$[i], $X_{d_x}$[j]\}
\State pairs = pairs + \{pair\_pos, pair\_neg\}
\State $Y_P$ = $Y_P$ +  [1, 0]
\EndFor
\EndFor
\State \Return pairs, $Y_P$
\EndFunction
\end{algorithmic} 
\end{algorithm} 

\clearpage

\begin{algorithm}[!h]
\caption{Create Reduced dataset \label{alg:reduce_dataset}} 
\begin{algorithmic}
\Function{ReduceDataset}{$X$, $Y_S$, $n_c$, NewSize}
\State  $m\gets$ int(NewSize/$n_c$) 
\State  indices=$[\,]$
\For{$d=1,2,\ldots,$ $n_c$}
\State class\_indices = where($Y_S$==d)
\State m\_class\_indices=random.choice(class\_indices, m)
\State indices.append(m\_class\_indices)
\EndFor
\State \Return $X$[indices], $Y_S$[indices]
\EndFunction
\end{algorithmic} 
\end{algorithm}

\begin{algorithm}[!h]
\caption{\emph{Sparse} Pair Label Noise (PLN)  \label{alg:sparse_PLN}}
   \begin{algorithmic}
       %\footnotesize
        \State $N \gets$ \text{number of pairs to train the SNN with}
        \State $\tilde{q} \gets$ \text{probability to apply transformation $\mc{T}_\text{PLN}$}
        \State $X,Y_S \gets$\text{load(dataset)}
        \State $n_c \gets$\text{len(unique($Y_S$))}
        \State pairs, $Y_P$ $\gets$ \textsc{CreateBalancedPairs}($X$, $Y_S$, $n_c$)
        \For{ $i= 1,2, \ldots, n_s$}
        \State indices=[]
        \State $Y_P\gets \mc{T}_\text{PLN}(\tilde{q},Y_P)\qquad $\hfill fraction of $Y_P$ gets randomized
        \State indices+=random.choice(where($Y_P$ ==0), N/2) $\qquad$ \hfill select balanced $N$ pairs
        \State indices+=random.choice(where($Y_P$ ==1), N/2)
        \State pairs$\gets$ pairs[indices]
        \State $Y_P\gets$ $Y_P$[indices]
        \State train
        \EndFor 
\end{algorithmic}
\end{algorithm}
\vspace{-0.1cm}
\begin{algorithm}[!h]
\caption{\emph{Sparse} Single Label Noise (SLN)\label{alg:sparse_SLN}}
   \begin{algorithmic}
        \State $N \gets$ \text{number of pairs to train the SNN with}
        \State $q \gets$ \text{probability to apply transformation $\mc{T}_\text{SLN}$}
        \State $X,Y_S \gets$\text{load(dataset)}
        \State $n_c \gets$\text{len(unique($Y_S$))}
        \For{ $i= 1,2, \ldots, n_s$}
        \State $Y_S'\gets \mc{T}_\text{SLN}(\tilde{q},Y_S)\qquad $  \hfill fraction of $Y_S$ gets randomized
        \State pairs, $Y_P$ $\gets$\textsc{CreateBalancedPairs}($X$, $Y_S'$, $n_c$)
        \State indices+=random.choice(where($Y_P$ ==0), N/2) $\qquad$ \hfill select balanced $N$ pairs 
        \State indices+=random.choice(where($Y_P$ ==1), N/2)
        \State pairs$\;\gets$ pairs[indices]
        \State $Y_P\gets$ $Y_P$[indices]
        \State train
        \EndFor 
    \end{algorithmic}
\end{algorithm}

\begin{algorithm}[!h]
\caption{\emph{Dense} Pair Label Noise  (PLN)\label{alg:dense_PLN}}
\begin{algorithmic}
        \State $N \gets$ \text{number of pairs to train the SNN with}
        \State $\tilde{q} \gets$ \text{probability to apply transformation $\mc{T}_P$}
        \State $X,Y_S \gets$\text{load(dataset)}
        \State $n_c \gets$\text{len(unique($Y_S$))}
        \For{ $i= 1,2, \ldots, n_s$}
        \State $X',Y_S' \gets$ \textsc{ReduceDataset}($X$, $Y_S$, $n_c$, N/2)
        \State pairs, $Y_P$ $\gets$\textsc{CreateBalancedPairs}($X'$, $Y_S'$, $n_c$)
        \State $Y_P\gets \mc{T}_\text{PLN}(\tilde{q},Y_P)\qquad $\hfill fraction of $Y_P$ gets randomized
        \State train
        \EndFor
    \end{algorithmic}
\end{algorithm}
\vspace{-0.15cm}
\begin{algorithm}[!h]

\caption{\emph{Dense} Single Label Noise (SLN)\label{alg:dense_SLN}}
\begin{algorithmic}
        \State $N \gets$ \text{number of pairs to train the SNN with}
        \State $q \gets$ \text{probability to apply transformation $\mc{T}_S$}
        \State $X,Y_S \gets$\text{load(dataset)}
        \State $n_c \gets$\text{len(unique($Y_S$))}
        \For{ $i= 1,2, \ldots, n_s$}
        \State $X',Y_S' \gets$ \textsc{ReduceDataset}($X$, $Y_S$, $n_c$, N/2)
        \State $Y_S'\gets \mc{T}_\text{SLN}(\tilde{q},Y_S')\qquad $\hfill fraction of $Y_S'$ gets randomized
        \State pairs, $Y_P$ $\gets$\textsc{CreateBalancedPairs}($X'$, $Y_S'$, $n_c$)
        \State train
        \EndFor 
    \end{algorithmic}
\end{algorithm}

\subsection{Effective noise derivation}
\label{app:prob}
We start by considering the
amount of effective noise (real mislabeling) introduced by the pair label transformation
\be
\mc{T}_{\text{PLN}}(\widetilde{q}):{y^P_{i}}\rightarrow \mbox{Random}(\{0,1\}) \qquad \mbox{ with
probability}\quad \widetilde{q}\,.
\ee
Each time we apply this transformation, the probability of a change in the pair label is 1/2, so the effective error probability is:
\be
P_{\text{PLN}} = \frac{\widetilde{q}}{2}\,.
\ee

This computation is slightly more complicated in the SLN case. Indeed, if we apply the following transformation 
\be
\mc{T}_{\text{SLN}}(q):{y_{i}}^{S} \rightarrow \mbox{Random}(\{1,\dots,n_c\})\qquad \mbox{ with
probability}\quad q\,,
\ee
on the initial dataset labels $y_{i}^S$, and we then create the pairs, the probability that one (and only one) element in a pair has been operated by $\mc{T}_{\text{SLN}}$ is
\be
P_{1L} = 2q(1 - q)\,,
\ee
while the probability that both elements have been operated by $\mc{T}_{\text{SLN}}$ is
\be
P_{2L} = q^{2}\,.
\ee
Now the question is: what is the probability that this single label
operation (we recall that the term \emph{single label} regards the application of $\mc{T}_{\text{SLN}}$ on the label of one or both pair elements and not on the
\emph{pair label}) leads to effective pair label corruption?
Let us assume that we have a pair of images belonging to different
classes $y^P=0$. The probability that the transformation of a single image label changes the pair label is equal to the likelihood that the same operation over both images effectively changes the pair label. The value of that probability is the following:
\be
{Q^{0 \rightarrow 1}}_{1L} = {Q^{0 \rightarrow 1}}_{2L} = \frac{1}{n_{c}}\,.
\ee
The same reasoning can be applied to pairs of objects belonging to the same class, $y^P=1$, and leads to
\be
{Q^{1 \rightarrow 0}}_{1L} = {Q^{1 \rightarrow 0}}_{2L} = \frac{(n_{c} - 1)}{n_{c}}\,.
\ee
Creating a balanced dataset where half of the pairs are equal and half are different is common practice. Therefore, we create a dataset where
\be
P_{y^P=1} = P_{y^P=0} = \frac{1}{2}\,.
\ee
Finally, we are now ready to estimate the amount of real noise that is introduced in our dataset corrupting single images labels. This is given by:
\be
\begin{array}{llll}
P_{\text{SLN}} &= P_{y^P=1}\ (P_{1L}{Q^{0 \rightarrow 1}}_{1L} + P_{2L}{Q^{0 \rightarrow 1}}_{2L})\\[5pt] &\qquad\qquad\qquad\qquad\qquad+ \ P_{y^P=0}(P_{1L\ }{Q^{1 \rightarrow 0}}_{1L} + P_{2L}{Q^{1 \rightarrow 0}}_{2L})\\[5pt]
&= \frac{1}{2} (P_{1L\ } + P_{2L})({Q^{0 \rightarrow 1}}_{1L} + {Q^{1 \rightarrow 0}}_{1L})\\[5pt]
&= q - \frac{1}{2}q^2\,.
\end{array}
\ee
Requiring that the effective dataset noise is the same in SLN and PLN setups, leads us to Eq.\,5 in the main text.
We want to stress that PLN and SLN  impose different constraints on the training process. PLN is a balanced noise source as the probability of transforming even pairs into odd ones, and vice versa is the same. On the other hand, SLN is an unbalanced source of noise, i.e., the probability that $\mc{T}_{\text{SLN}}$ transforms equal pairs into different ones, $(n_c-1)/n_c$, is in general much higher than the opposite case, $1/n_c$.
Moreover, as opposed to classification tasks, in Siamese networks and contrastive learning, noise can generally lead to inconsistent relations in the training set. 
A similarity relation is defined by reflexivity, symmetry, and transitivity, but the appearance of noise can compromise this last property. In fact, PLN, randomly shuffling pair labels, leads to inconsistent relations in the dataset. This effect becomes more apparent as we increase the density (number of links) in our training set. On the other hand, similarity breaking does not appear in SLN, where the similarity relations may go against image features but are always self-consistent.

% \newpage 

\subsection{Output layer and loss function}
\label{app:distances_and_losses}

We perform our experiments using the two different output layers and loss functions described below.

%\vspace{-0.3cm}

\paragraph{Euclidean distance and Contrastive Loss.} In this first case, the Siamese NN output layer computes the Euclidean distance between the output vectors of the Siamese branches, $\vec{z}(x)$. Therefore, the model prediction that quantifies the similarity between the two images in a pair is given by:
\begin{equation}
d_i=||\vec{z}(x_1^i)-\vec{z}(x_2^i)||_2\,.
\end{equation}
We then train the network considering the contrastive loss function\footnote{R. Hadsell, S. Chopra, Y. LeCun,
  2006 IEEE Computer Society Conference on Computer Vision and Pattern Recognition, \emph{Dimensionality Reduction by Learning an Invariant Mapping}.}:
\begin{equation}
%\small
\label{eq:contrastive_loss}
\begin{array}{ll}
\mathcal{L}(y^P, d) =  \displaystyle\frac{1}{N_\text{pairs}} \sum_i \Big[ y_i^P \, d_i^2  \left. + (1- y_i^P) \left[\text{max}(0, m -d_i)\right]^2\right],\\
\end{array}
\end{equation}
\noindent where $y^P_i$ is the true label and $m$ sets the threshold at which the network classifies a given pair as similar or different. During  training, the network tries to minimize $\mathcal{L}$ by collapsing similar samples and pulling apart  different samples by a distance equal to the margin, $m$. The accuracy is given by: 
\begin{eqnarray}
\mbox{acc}=1-\mbox{err}=1-\frac{1}{2N}\sum_i |y^P_i - \hat{y}(d_i)|\,,   \quad \mbox{where} \quad  \hat{y}(d_i)=[\mathbbm{1}_{d < m/2}](d_i).
\end{eqnarray}
In all experiments, we choose the margin to be $m=1$.

\paragraph{Cosine similarity and Cosine Embedding Loss.} In this setup, the output layer computes the cosine similarity between the output vectors of the Siamese branches. The model prediction is thus given by:
\begin{equation}
s_i=\displaystyle \cos\left(\frac{\vec{z}(x_1^i)\cdot\vec{z}(x_2^i)}{||\vec{z}(x_1^i)||_2 ||\vec{z}(x_2^i)||_2}\right)\,.
\end{equation}
We train the network using the Cosine Embedding Loss function, 
\begin{equation}
\label{eq:cosine_loss}
\begin{array}{ll}
\mathcal{L}(y^P, s) =  \displaystyle\frac{1}{N_\text{pairs}} \sum_i \Big[ y_i^P \, (1-s_i)  
\left. + \,(1- y_i^P)\, \text{max}(0, s_i -\cos(\alpha))\right],\\
\end{array}
\end{equation}
according to which similar images should give rise to vectors pointing in roughly the same direction. In contrast, the angle between vectors coming from different images should be larger than or equal to $\alpha$. In this model, we compute the accuracy as:
\begin{eqnarray}
\mbox{acc}=1-\frac{1}{2N}\sum_i |y^P_i - \hat{y}(s_i)|, \quad \mbox{where} \quad \hat{y}(s_i)=[\mathbbm{1}_{s \,> \cos(\alpha/2)}](s_i).
\end{eqnarray}
In all experiments, we choose $\alpha=\pi/3$.

\end{document}